\documentclass[sigconf]{acmart}

\usepackage{booktabs} 
\usepackage{multirow}
\usepackage{hhline}
\usepackage{tabu}

\setcopyright{none}

\DeclareMathOperator*{\med}{median} 

\graphicspath{{Images/}}

\begin{document}
\title{Mining Automatically Estimated Poses from Video Recordings of Top Athletes}

\author{Rainer Lienhart}
\authornote{This work was performed during his sabbatical at FXPAL, Palo Alto, CA, USA}
\affiliation{%
  \institution{University of Augsburg}
  \streetaddress{Universitaetsstr. 6a}
  \postcode{86159}
}
\email{rainer.lienhart@informatik.uni-augsburg.de}

\author{Moritz Einfalt}
\affiliation{%
  \institution{University of Augsburg}
  \streetaddress{Universitaetsstr. 6a}
  \postcode{86159}
}
\email{moritz.einfalt@informatik.uni-augsburg.de}

\author{Dan Zecha}
\affiliation{%
  \institution{University of Augsburg}
  \streetaddress{Universitaetsstr. 6a}
  \postcode{86159}
}
\email{dan.zecha@informatik.uni-augsburg.de}



\begin{abstract}
  Human pose detection systems based on state-of-the-art DNNs are on the go to be extended, adapted and
  re-trained to fit the application domain of specific sports. Therefore, plenty of noisy pose data will
  soon be available from videos recorded at a regular and frequent basis. This work is among the first to
  develop mining algorithms that can mine the expected abundance of noisy and annotation-free pose data
  from video recordings in individual sports. Using swimming as an example of a sport with dominant
  cyclic motion, we show how to determine unsupervised time-continuous cycle speeds and temporally
  striking poses as well as measure unsupervised cycle stability over time. Additionally, we use
  long jump as an example of a sport with a rigid phase-based motion to present a technique to
  automatically partition the temporally estimated pose sequences into their respective phases.
  This enables the extraction of performance relevant, pose-based metrics currently used by
  national professional sports associations. Experimental results prove the effectiveness of our
  mining algorithms, which can also be applied to other cycle-based or phase-based types of sport. 
\end{abstract}


%
%



\settopmatter{printacmref=False}

\maketitle


\section{Introduction}
\label{sec:1}

Since the arrival of deep neural networks (DNNs), state-of-the-art DNN-based human pose estimation systems
have made huge progress in detection performance and precision on benchmark datasets
\cite{Wei2016, Andriluka2014, Chu2017, Yang2017, Newell2016}. Recently, these
research systems have been extended, adapted and re-trained to fit the application domain of specific
sports \cite{Zecha2017, Einfalt2018}.
Soon they will disrupt current performance analyses in all kinds of sport as the amount of
available pose data will explode due to automation. So far, pose detection and analysis of top-class
athletes has been very time-consuming manual work. It was scarcely performed by the national professional
sports associations for them and almost never for athletes below that level. The forthcoming availability
of automatic pose detection systems will make plenty of noisy pose data available from videos recorded at
a much more regular and frequent basis. Despite this imminent change in data quantity at the cost of
probably higher noise in the pose data, very little research has been devoted to explore the opportunities
of extracting informative and performance relevant information from these pose detection results
through data mining. This work is focusing on this question and presents a set of unsupervised pose
mining algorithms that extract or enable extraction of important information about athletes and how
they compare to their peers. We will use world-class swimmers in the swimming channels as an example
of a sport with dominant cyclical motion and long jumping as an example of a sport with clear
chronologically sequential phases.

\begin{figure}[tbh]
  \centering
  \includegraphics[width=0.96\columnwidth]{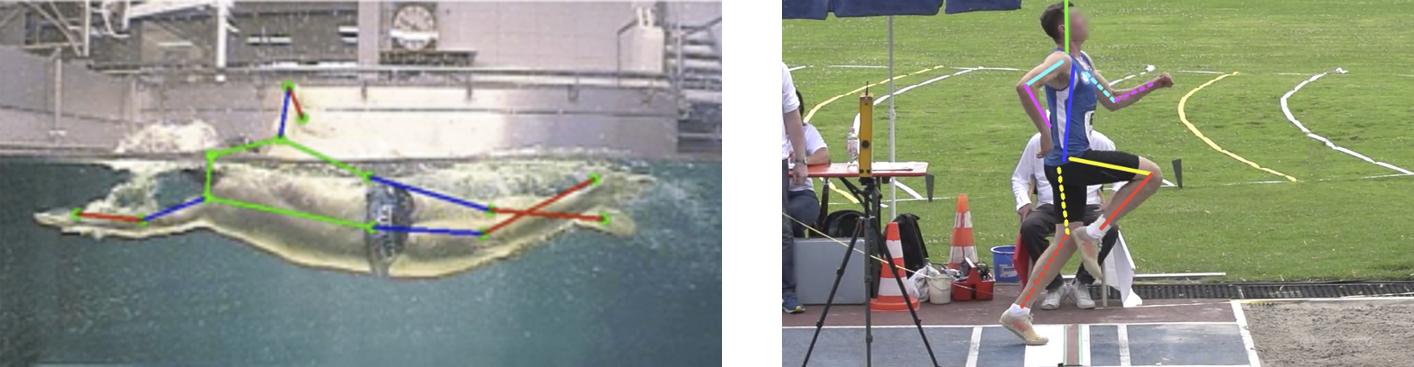}
  \caption{Detected poses of a swimmer and a long jumper.}
  \label{fig:pose_examples}
\end{figure}

In this work, pose data denotes the noisy poses produced by some image or video-based pose detection system, either with or without customized post-processing to identify and clean out errors by interpolation and/or smoothing. Our pose data is based on the image-based pose detection system presented in \cite{Einfalt2018} and \cite{Wei2016}. Examples are depicted in Figure~\ref{fig:pose_examples}.

\textbf{Contributions:} (1) Our research work is among the first that does not mine manually annotated
poses with little noise (because of manual annotations by professional coaches and support staff),
but rather focus on the noisy output of a DNN-based pose detection system lacking
\textbf{any} pose annotations. (2) All manual annotations are typically confined to a few key poses
during the relevant actions (i.e., they are temporally sparse), and so are the derived key performance
parameters. We, however, exploit that pose detection systems can process every frame, producing a
temporally dense output by robustly estimating the performance parameters time-continuously at every
frame. (3) Some sports are dominated by cyclical motion, some by clear chronologically sequential phases.
We present our mining algorithms to extract or to enable extraction of key performance parameters by
picking swimming as a representative of a cyclical kind of sport and long jumping as one of the second
type of sport.


\section{Related Work}
\label{sec:2}

Human pose based semantic data mining research is dominated by works on motion segmentation and
clustering, key-pose identification and action recognition. While dimensionality and representation of
poses may differ across recent works, the goal often is to allow for retrieval and indexing of
human pose/motion in large video databases or classification of motion sequences at different abstraction levels.

\textbf{Human pose mining:} Both works in \cite{Ren2011} and\cite{Voegele2014}
cluster 3D motion capture data and determine algorithmically similar motion sequences for
database retrieval, while \cite{Sedmidubsky2013} develops a similarity algorithm for comparing key-poses,
subsequently allowing for indexing motion features in human motion databases. For the task of
action recognition, \cite{Lv2007} and \cite{Baysal2010} perform clustering on shape based representations
of 2d human poses and learn weights to favor distinctive key-poses. Both show that temporal context
is superfluous if human poses with high discriminative power are used for action recognition.
Data mining for action recognition based solely on joint location estimates is still scarce.
\cite{Wang2013} propose spatial-part-sets obtained from clustering parts of the human pose to obtain
distinctive, co-occurring spatial configurations of body parts. They show that these sets improve the
task of action recognition and additionally the initial pose estimates.

\textbf{Pose mining in sports:} In the field of sport footage analysis, the task of action
recognition often translates to the identification of specific motion sequences within a sport activity.
\cite{DeSouza2016} use latent-dynamic conditional random fields on RGB-d skeleton estimates of Taekwondo
fighters to identify specific kicks and punches in a fight sequence. Long jump video indexing has been
researched by \cite{Wu2002}, who perform motion estimation and segmentation of camera and athlete motion
velocity to extract and classify semantic sequences of long jump athletes. \cite{Li2010} build a similar
system for high diving athletes. They also derive human pose from shape and train a Hidden Markov Model
to classify a partial motion of jumps.

The extraction of kinematic parameters of athletes from video footage, specifically stroke rates of
swimmers, was recently researched by \cite{Victor2017}, who perform stroke frequency detection on
athletes in a generic swimming pool. \cite{Zecha2017} derive additional kinematic parameters from
swimmers in a swimming channel by determining inner-cyclic interval lengths and frequencies through
key-pose retrieval. Compared to other approaches that rely on the concept of identifying key-poses,
their approach lets a human expert define what a discriminative key-pose should be.

\textbf{Our work:} While our work is influenced by the related work above, the major difference is
that we only use raw joint estimates from a human pose estimator while previous work heavily relies
either on correctly annotated ground truth data to train models or recordings from motion capture
RGB-d systems. Additionally, our work connects data mining on human pose estimates with the extraction
of kinematic parameters of top athletes.


\section{Measuring Pose Similarity}
\label{sec:3}

In computer vision, the human pose at a given time is defined by a set of locations of important
key points on a human such as joint locations. The number of key points varies based on the application
domain. In the analysis of top-level athletes, the pose is the basis of many key performance indicators
and may also include points on the device(s) the athlete is using. Since the pose is so central to most
sports-related performance indicators, we need to be able to reliably evaluate the similarity or
distance between poses. This section develops our metric pose distance measure that is invariant to
translation, scale and rotation in the image plane. It will be used in all algorithms discussed in
Sections \ref{sec:4} to \ref{sec:5}.

Throughout the paper, we assume that all video recordings have been processed by some
pose detection system. In our case, we use the system from \cite{Einfalt2018} for swimming
and \cite{Wei2016} for long jump. We do not expect to have a pose for all frames.
Through some parts of a video, the athlete might not be completely in the picture, if present at all.
Or the detection conditions are so difficult that the detection system does not detect any pose.
Our mining algorithms have to deal with that. However, we discard all poses that are only partially
detected to make mining simpler.

\subsection{Pose}
\label{sec:3.1pose}

Mathematically, a 2D pose $p$  is nothing but a sequence of $N$ two-dimensional points,
where each 2D point by convention specifies the coordinates of the center of a joint location
or of some other reference location on the human or object(s) under investigation:
\begin{equation}
  \label{eq:pose_def}
  p = \left\{  \left( x_k, y_k  \right) \right\}^{N}_{k=1} \equiv \begin{pmatrix} x_1 & \dotsb  & x_N \\ y_1 & \dotsb  & y_N \end{pmatrix}
\end{equation}
Our human pose model consists of $N=14$  joints. Throughout the paper, a  \textit{pose clip} and \textit{pose sequence} denote a temporal sequence $\mathbf{p}_{t1:t2}$ of poses $[ p_{t1}, p_{t1+1}, \dotsc, p_{t2-1}, p_{t2} ]$. The term \textit{pose clip} hints at a short temporal pose sequences (e.g. $\frac{1}{2}$ to $2$ seconds), while \textit{pose sequence} often refers to much longer durations -- up to the complete video duration (e.g., $30$ seconds and longer). Video time and time intervals are usually expressed using sequential frame numbers as we assume recordings at a constant frame rate.

\subsection{Aligning Two Poses}
\label{sec:3.2aligning}

Before we can define our pose distance measure, we need to specify how we align a pose $p$ to a given reference pose $p_r$ by finding the scaling factor $s$ , rotation angle $\theta$ and translation $t=(t_x, t_y)$, which applied to each joint of $p$ results in $p^\prime$, which minimizes the mean square error (MSE) between the transformed pose $p^\prime$ and the reference pose $p_r$ \cite{Rowley1998}:
\begin{equation}
  \label{eq:mse}
  MSE(p_r, p) : = MSE(p_r, p^\prime) = \frac{1}{2N} \lVert p_{r, reshaped} - p^\prime_{reshaped}  \rVert^{2}_{2}
\end{equation}
with
\begin{equation}
  \label{eq:ttrans}
  t_{trans} = (a, b, t_x, t_y)^T   
\end{equation}
and
\begin{equation}
  \label{eq:p_matrix}
    p^\prime_{reshaped}  := \begin{pmatrix} x^\prime_1 \\ y^\prime_1 \\ x^\prime_2 \\ y^\prime_2  \end{pmatrix} = \begin{pmatrix} x_1 & -y_1 & 1 & 0 \\ y_1 & x_1 & 0 & 1 \\ x_2 & -y_2 & 1 & 0 \\ y_2 & x_2 & 0 & 1 \\  \end{pmatrix} \begin{pmatrix} a \\ b \\ t_x \\ t_y  \end{pmatrix}  =: A \cdot t_{trans}
\end{equation}
Note that the $N \times 2$  matrix $p^\prime$  is reshaped to a $2N \times 1$  vector $p^\prime_{reshaped}$. The pseudo-inverse $t^{opt}_{trans} = (A^TA)^{-1}A^Tp_{r, reshaped}$ gives us in closed form the transformation of pose $p$ that minimizes the mean squared error between the joints of reference pose $p_r$ and transformed pose $p^\prime$. Each joint $(x,y)$  of $p$ is mapped to
\begin{equation}
  \label{eq:pose_transform}
  \begin{pmatrix} x^{\prime} \\ y^{\prime} \end{pmatrix} = \begin{pmatrix} s \cos \theta & -s \sin \theta \\ s \sin \theta & s \cos \theta \end{pmatrix} \begin{pmatrix} x \\ y \end{pmatrix} + \begin{pmatrix} t_x \\ t_y \end{pmatrix} = \begin{pmatrix} a & -b & t_x \\ b & a & t_y \end{pmatrix} \begin{pmatrix} x \\ y \\ 1 \end{pmatrix}
\end{equation}
using the optimal transformation $t^{opt}_{trans}$. The associated $MSE$ value indicates how well a pose fits a reference pose. Thus, given a set of poses, their associated $MSE$ values can be used to rank these poses according to their fitness to the reference pose. However, two peculiarities about $MSE(p_r, p)$  need to be emphasized:
\begin{enumerate}
\item It is not symmetric, i.e., generally $MSE(p_r, p) \neq MSE(p, p_r)$. The reason for this is that the pose is always scaled to the size of the reference pose. Thus, if their two scales are very different, so will be $MSE(p_r, p)$ and $MSE(p, p_r)$.
\item Its magnitude depends on the scale of the reference pose. Doubling the reference pose's scale will quadruple the $MSE$  value. Thus, if a pose is compared against various reference poses, the scale of the references poses matters.
\end{enumerate}
Both peculiarities of the $MSE(p_r, p)$ value suggest that we need to normalize the poses we are comparing to get universally comparable MSE values and thus a universally applicable distance measure between two poses.

\subsection{Pose Distance Measure}
\label{sec:3.3pose_distance_measure}
It is common in pose detection evaluation to scale a reference pose by assigning a fixed size either to the length of the distance between two characteristic points of the pose or to the head. While using a single rectangle or two reference points may be fine in case of ground truth annotations, it is statistically not advisable for noisy detection results. We need a normalization that is based on more joints to reduce noise. Hence the scale $s_p$ of pose $p$ is defined as the average distance of all joints of a pose to its center of mass $c_p = (c_{p,x}, c_{p,y})^T$:
\begin{equation}
  \label{eq:pose_scale}
  s_p = \frac{1}{N} \sum_{k=1}^{N} \begin{Vmatrix} \begin{pmatrix} x_k \\ y_k \end{pmatrix} - \begin{pmatrix} c_{p,x} \\ c_{p,y} \end{pmatrix} \end{Vmatrix}_2
\end{equation}
with
\begin{equation}
  \label{eq:center_of_mass}
  \begin{pmatrix} c_{p,x} \\ c_{p,y} \end{pmatrix} = \frac{1}{N} \sum_{k=1}^{N} \begin{pmatrix} x_k \\ y_k \end{pmatrix}
\end{equation}
Given an arbitrary reference scale $s_{ref}$, we define our symmetric translation, rotation and scale invariant distance measure between two poses as
\begin{equation}
  \label{eq:mse_norm}
  MSE_{norm}(p_1, p_2) = \frac{s^2_{ref}}{2s^2_{p_1}} MSE(p_1, p_2) + \frac{s^2_{ref}}{2s^2_{p_2}} MSE(p_2, p_1)
\end{equation}
It enables us to judge pose similarity between poses derived from videos recorded by different cameras, at different locations and distances to the athletes.


\section{Mining Pose Data of Swimmers}
\label{sec:4}

Cyclical motions play a decisive and dominant role in numerous sports disciplines, e.g., in cycling, rowing, running, and swimming. In this section, we use swimming as an example to explore what kind of automated mining we can perform on the detected noisy poses. We use the pose data derived from world class swimmers recorded in a swimming channel. A single athlete jumps into the flowing water against the flow (from the right in Figure~\ref{fig:pose_examples} left), swims to the middle in any manner (e.g., by an extended set of underwater kicks or by freestyle on the water surface) and then starts the cyclic stroke under test. The video recording can start any time between the dive and the action of interest (= swimming a stroke) and stops shortly after it ended. During most of the recording time the athlete executes the cyclic motion under test.

\subsection{Time-Continuous Cycle Speeds}
\label{sec:4.1cycle_speeds}

For all types of sport with dominant cyclical motions, the change in cycle speed over time is a very indicative performance parameter. It can be derived through data mining without providing any knowledge to the system, but the automatically detected joint locations for each pose throughout a video sequence. Given a pose at time $t$, the \textit{cycle speed} at time $t$ is defined as $1$ over the time needed to arrive at this pose from the same pose one cycle before. In the case of a swimmer, the desired cycle speed information is strokes per minutes, which can be derived from the stroke length in frames given the video sampling rate in frames per seconds by
\begin{equation}
  \label{eq:stroke_rate}
  \frac{\text{\# strokes}}{\text{minute}} = \left(\frac{\text{\# frames}}{\text{stroke}}\right)^{-1} \cdot \frac{\text{\# frames}}{\text{seconds}} \cdot \frac{60 \text{ seconds}}{\text{minute}}
\end{equation}
The \textit{stroke length} is measured by the number of frames passed from the same pose one cycle before to the current pose.

In the following, we describe the individual steps of our statistically robust algorithm to extract time-continuous cycle speeds by first stating the characteristic property of cyclic motion we exploit, followed by an explanation how we exploit it. The adjective \textit{time-continuous} denotes that we will estimate the \textbf{cycle speed for every frame} of a video in which the cyclic motion is performed:
\begin{enumerate}
\item \textbf{Input:} A sequence $P$ of poses $p$ for a video: $P=\{(f_p, p)\}_{f_p}$.\\
  It contains pairs consisting of a detected pose $p$ and a frame number $f_p$ in which it was detected. The subscript $f$ in $\{(f,\dotsc)\}$ indicates that the elements in the set $\{\dotsc\}$ are ordered and indexed by frame number $f$. Note that we might not have a pose for every video frame.
\item \textbf{Property:} Different phases of a cycle and their associated poses are run through regularly. As a consequence a pose $p$ from a cycle should match periodically at cycle speed with poses in $P$. These matching poses $p^\prime$ to a given pose $p$ identify themselves visually as minima in the graph plotting the frame number of poses $p^\prime$ against its normalized distance to given pose $p$. Therefore, we compare every pose  $p$ in a video against every other pose $p^\prime$ and keep for each pose $p$ a list $L_p$ of matches:
  \begin{equation}
    \label{eq:match_list}
    L_p = \left\{ \left( f_{p^\prime}, p^\prime, MSE_{norm}\left(p, p^\prime\right) \right) \right\}_{f_{p^\prime}} \;\;\; \forall p \in P
  \end{equation}
  Poses match if their normalized $MSE$ value is below a given threshold. For a target scale of $s_{ref}=100$ we use a threshold of $49$ (on avg. $7$ pixels in each direction for each joint).
\item \textbf{Property:} Not every pose is temporally striking.\\
  An athlete might stay for some time even during a cycle in a very similar pose, e.g., in streamline position in breaststroke after bringing the arms forward. However, at one point this specific pose will end to enter the next phase of the cycle. Thus, from step 2, we sometime not only get the correct matches, but also nearby close matches. We consolidate our raw matches in $L_p$ by first temporally clustering poses $p^\prime$. A new cluster is started if a gap of more than a few frames lies between two chronologically consecutive poses in $L_p$. Each temporal cluster is then consolidated to the pose $p_c$  with minimal normalized $MSE$ to the pose $p$. The cluster is also attributed with its \textit{temporal spread}, i.e., the maximal temporal distance of a pose in the cluster from the frame with the consolidated pose $p_c$, leading us to the \textit{reoccurrence sequences} $L^\prime_p$ with
  \begin{equation}
    \label{eq:reoccurence_seq}
    L^\prime_p = \left\{ \left( f_{p_c}, p_c, spread \right)  \right\}_{f_{p}} \;\;\; \forall p \in P
  \end{equation}
  and for the complete video to $L_{video}=\left\{ \left( f_p, p, L^\prime_p \right) \right\}_{f_{p}}$.
\item \textbf{Property:} Temporally non-striking poses are unsuitable to identify cyclic motion. Therefore, all clusters with a temporal spread larger than a given threshold are deleted.\\
  In our experiments we set this value to $10$ frames, resulting in
  \begin{equation}
    \label{eq:striking_poses}
    L^{\prime \prime}_p = \left \{ \left( f_{p_c}, p_c, spread \right) \middle \vert spread < 10 \right \}_{f_p} \;\;\; \forall p \in P.
  \end{equation}
\item \textbf{Property:} Most of the time the video shows the athlete executing the cyclical motion under test. Consequently, poses from the cyclic motion should most often be found. \\
  Hence, we create a histogram over the lengths of the reoccurrences sequences $(\equiv \vert L^{\prime \prime}_p \vert)$ for the various poses $p$. We decided to keep only those reoccurrence sequences $L^{\prime \prime}_p$ which belong to the $50\%$ longest ones:
  \begin{equation}
    \label{eq:longest_sequences}
    L^{\prime}_{video} = \left\{ \left( f_p, p, L^{\prime \prime}_p \right) \middle \vert \left \vert L^{\prime \prime}_p \right \vert \geq \med_{p \in P} \left( \left \vert L^{\prime \prime}_p \right \vert \right) \right\}_{f_p}
  \end{equation}
\item \textbf{Property:} The observed difference of the frame numbers in each reoccurrence sequence in $L^{\prime}_{video}$ between two chronologically consecutive matches should most frequently reflect the actual stroke length.\\
Figure~\ref{fig:reoccurence_sequences} shows two sample plots. On the x-axis, we have the minuend of the difference and the difference value on the y-axis. The blue and yellow dots display all observed difference values from $L^{\prime}_{video}$. From them we derive our final robust estimate by local median filtering in two steps: (1) We take each frame number $f$ with at least one difference value and determine the median of the observed stroke lengths (= difference values) in a window of $\pm 2$ seconds (approx. $2$ to $4$ stroke cycles). We remove all difference values at frame number $f$, which deviate more than $10\%$ from the median. E.g., @$50$ fps a median stroke length of $60$ frames results in keeping only difference values in $[54,66]$. The deleted difference values are shown in yellow in Figure~\ref{fig:reoccurence_sequences}, while the remaining ones are shown in blue. (2) We piecewise approximate the remaining data points with a polynomial of degree $5$ over roughly $3$ cycles while simultaneously enforcing a smoothness condition at the piecewise boundaries.
  \begin{figure}[tb]
    \centering
    \includegraphics[width=0.98\columnwidth]{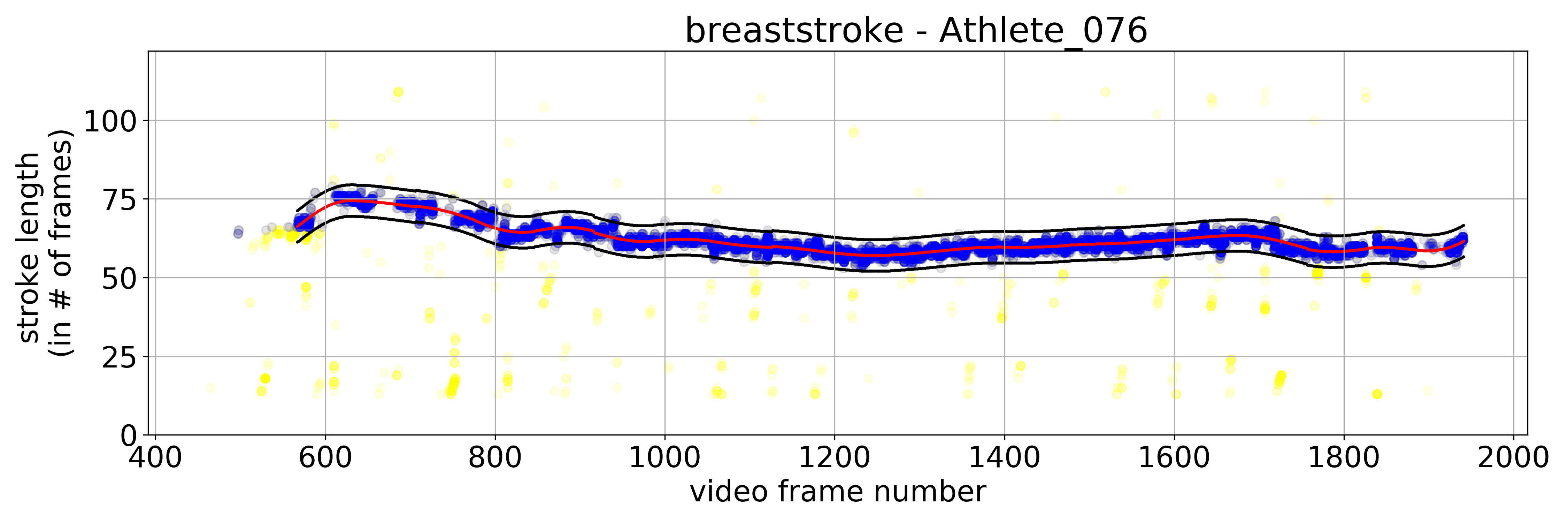}\\
    \includegraphics[width=0.98\columnwidth]{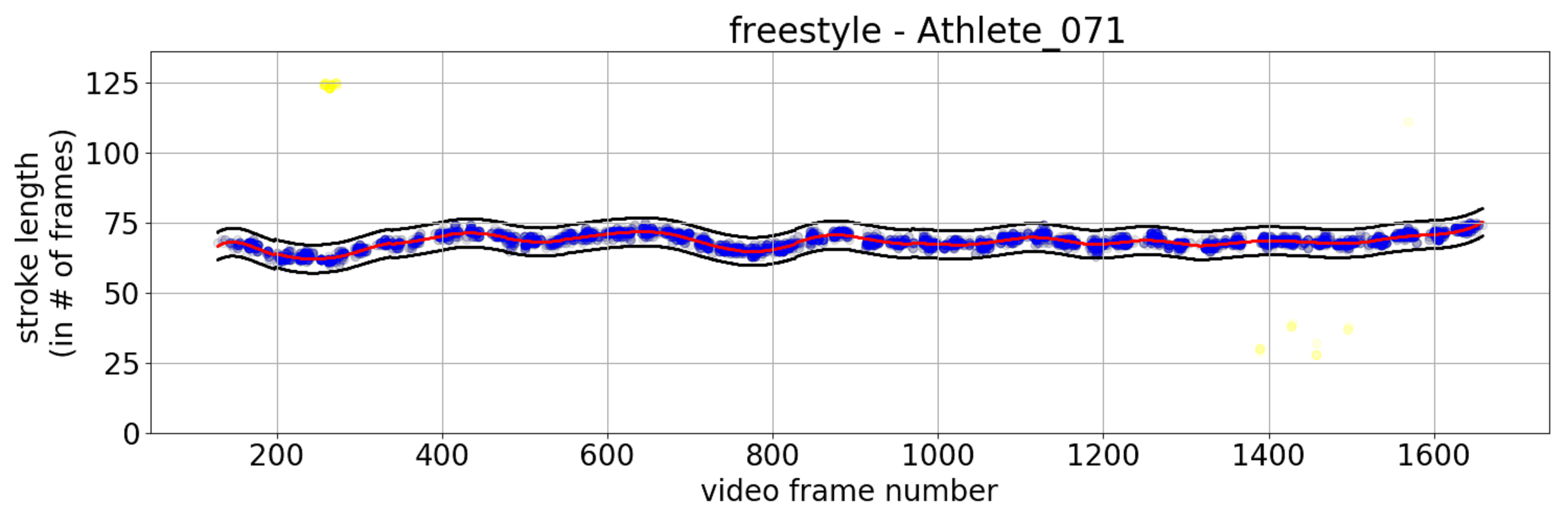}
    \caption{Examples showing frame differences between chronologically consecutive matches of  in all reoccurrence sequences of   against frame number. The red line visualizes the time-continuous estimate of stroke cycle length, with black lines indicated the $\pm 10\%$ corridor.}
    \label{fig:reoccurence_sequences}
  \end{figure}
\end{enumerate}
This approximation gives us our time-continuous estimates of the stroke cycle length over the interval in the video throughout which the stroke was performed. As a side effect it also automatically identifies the temporal range in the video during which the stroke was performed by the frame number ranges for which we have cycle speeds. The same technique is applicable to determine the kicks per minutes for freestyle and backstroke by restricting the pose to joints from the hip downwards.

\subsection{Temporally Striking Poses}
\label{sec:4.2temp_striking_poses}

During a cyclical motion some poses are more striking than others with respect to a given criterion. One such highly relevant criterion is how well a repeating pose can be localized temporally, i.e., how unique and salient it is with respect to its temporally nearby poses. The temporally most striking poses can be used, e.g., to align multiple cycles of the same swimmer for visual comparison.

Commonly, local salience is measured by comparing the local reference to its surrounding. In our case the local reference is a pose $p_r$ at frame $r$ or a short sequence of poses $p_{r-\triangle w_l}, \dotsc, p_r, \dotsc, p_{r+\triangle w_l}$  centered around that pose, and we compare the sequence to the temporally nearby poses. Thus, we can compute saliency by:
\begin{equation*}
  \label{eq:saliency}
  saliency \left( p_r \right) = \sum_{\triangle w_s = -w_s}^{w_s}  \sum_{\triangle w_l= -w_l}^{w_l}
  \frac{MSE \left( p_{r + \triangle w_l}, p_{r + \triangle w_l + \triangle w_s} \right)} {\left( 2 w_s + 1  \right) \left( 2 w_l + 1 \right)}
\end{equation*}
Experimentally, the saliency measure was insensitive with respect to the choices of $w_l$ and $w_s$. Both were arbitrarily set to $4$.

The salience values for each pose during the cyclic motion of a video can be exploited to extract the $K$ most salient poses of a cycle. Hereto, we take the top $N$ most salient poses ($N \gg K$) and cluster them with affinity propagation (AP) \cite{Frey2007}. Salient poses due to pose errors will be in small clusters, while our most representative poses are the representative poses of the $K$ largest clusters.

\begin{figure}[tb]
  \centering
  \includegraphics[width=0.98\columnwidth]{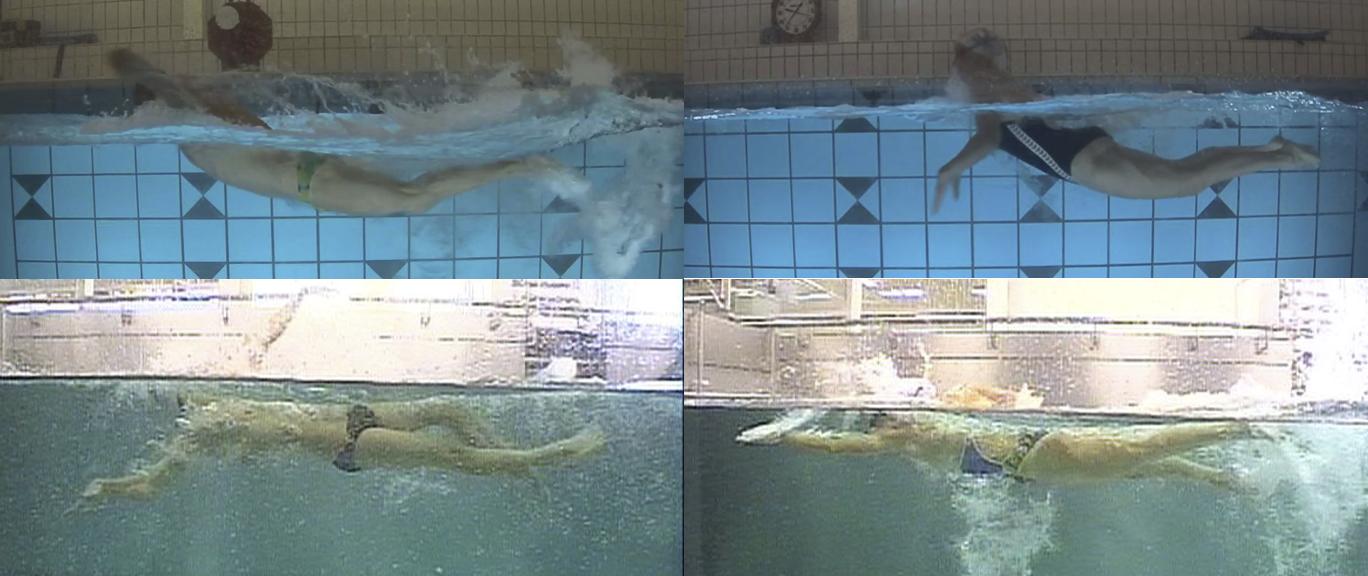}
  \caption{Examples of temporally striking poses; top left to bottom right: fly, breast, back and free.}
  \label{fig:striking_poses}
\end{figure}

For determining the most salient pose of an athlete's stroke, it is sufficient to pick the top $20$ most salient poses, cluster them with AP and retrieve the cluster representative with the most poses assigned. Figure~\ref{fig:striking_poses} shows one example for each stroke. Note that the most salient pose is another mean to determine the cycle speed reliably cycle by cycle, as this pose is most reliably localized in time. However, we only get one cycle speed value per cycle. 

\subsection{Cycle Stability}
\label{sec:4.3cycle_stability}

A common and decisive feature among winning top athletes is their trait to show off a very stable stroke pattern over time, under increasing fatigue and at different pace. One way to measure stroke cycle stability is to select a reference pose clip of one complete cycle and match this reference pose clip repeatedly over the complete pose sequence of the same video or a set of pose sequences derived from a set of videos recordings of some performance test (e.g., the $5 \times 200m$ step test after Pansold \cite{Pyne2001, Pansold1985}). Given all these clip matches and their associated matching scores, an average score of matching can be computed and taken as an indicator of stroke cycle stability: The better the average matching score, the more stable the stroke of the athlete. Alternatively, the matching score may be plotted versus time in order to analyze, how much the stroke changes from the desired one over (race) time. A reference pose cycle may automatically be chosen by selecting a clip between two contiguous occurrences of a temporally striking pose or by specifying a desired/ideal stroke cycle.

\textbf{Levenshtein distance:} With regards to that goal, we first turn our attention to the task of how to match a pose clip to a longer pose sequence and compute matching scores. We phrase the task to solve in terms of the well-studied problem of approximate substring matching: The task of finding all matches of a substring $pat$ in a longer document $text$, while allowing up to some specified level of discrepancies. In our application, a pose represents a character and a clip/sequence of poses our substring/document. The difference between `characters' is measured by a $[0,1]$-bounded distance function derived from the normalized $MSE$ between two poses:
\begin{align*}
  \label{eq:lev_pose_distance}
  & dist\_fct \left( p_1, p_2 \right) = \nonumber \\
  & = \begin{cases}
    0& \text{ if } MSE_n\left( p_1, p_2 \right) \leq th_{same}\\
    \frac{MSE_n\left( p_1, p_2 \right) - th_{same}}{th_{diff} - th_{same}}& \text{ if } MSE_n\left( p_1, p_2 \right) \geq th_{diff} \\
    1& \text{ else.} \\
\end{cases}
\end{align*}

The cost of transforming one pose into another is $0$ for poses which are considered the same ($MSE_n( p_1, p_2 ) \leq th_{same}$) and 1 for poses which are considered different ($MSE_n( p_1, p_2 ) \geq th_{diff}$). Between these two extremes, the transformation cost is linearly scaled based on the $MSE_n$ value. 

Any algorithm to compute the Levenshtein distance \cite{Levenshtein1966, Meyers1994} and its generalization called edit distance is suitable to perform matching and compute a matching score between a search pattern $pat$ and a longer document $text$ at every possible end point location of a match within $text$. It results in a matrix $d$ of matching costs of size $len(pat) \times len(text)$, where $d[i,j]$ is the cost of matching the first $i$ characters of $pat$ up to end point $j$ in $text$.

We use our custom distance function not only for transformations, but also for insertions and deletions. We deliberately made this chose as it better fits the characteristic of swimming: The absolute duration of a stroke cycle, i.e. the number of poses in a sequence, depends on the pace of the swimmer. However, the better the athlete, the more consistent he/she executes the pose successions across different paces. We therefore do not want to see an additional cost if, e.g., a swimmer stays longer/shorter in a perfect streamline position or if he/she goes slower/faster through the recovery phase of a stroke cycle than the reference clip. Pace is already captured by the cycle speed. Here we only want to focus on the stability of the stroke pattern, no matter how fast the stroke is executed. Note that swimmers with less than perfect swimming technique typically modify their poses when changing pace.

\begin{figure}[tb]
  \centering
  \includegraphics[width=0.7\columnwidth]{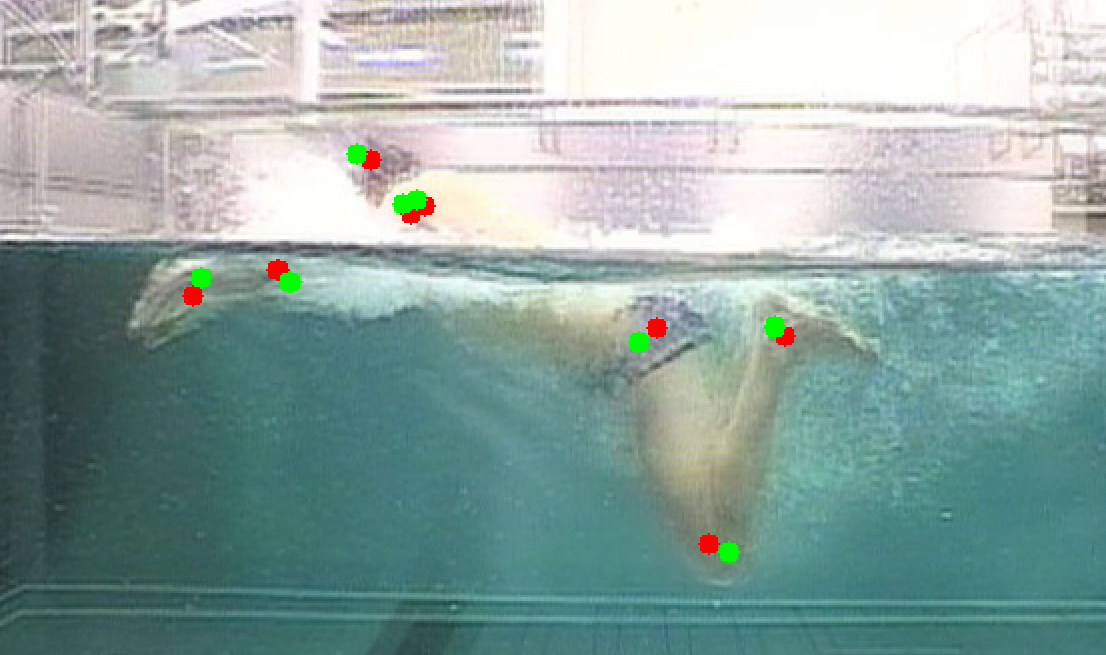}
  \caption{Alignment example of the same swimmer at different stroke cycles. Joints of the reference/matching pose are in shown in red/ green.}
  \label{fig:pose_alignment}
\end{figure}

\textbf{Match extraction:} The matching distances $d[len(pat),j]$ of the complete search pattern $pat$ computed by the edit distance at end point  $j$ in $text$ are normalized by the virtual matching length, i.e., by the number of transformations, deletions and insertions needed for that match. We call this $len(text)$-dimensional vector of normalized matching scores over all possible end points in $text$ $score_{match}(pat, text)$.  All clear minima in it identify the end points of all matches of the pose clip to the sequence together with the associated matching distances. Since our pose clips are highly specific in matching, our minima search does not require any non-maximum suppression. The matching sequence is derived by backtracking from this end point to the beginning of the match by using $d[i,j]$. Figure~\ref{fig:pose_alignment} found shows one example of matched poses of two different stroke cycles.

\textbf{Athlete Recognition:} While we were matching a given pose clip to all videos in our video database, we accidentally discovered that $score_{match}$ is also a perfect tool to automatically recognize a specific athlete. Usually, when matching a pose clip to the pose sequence of a different male or female swimmer, $score_{match}$ is 4 to 8 times higher in comparison to the score computed against the video the pose clip was taken from. However, in this case the matching score was as low as matched against the same video despite being a recording at a different test in a different swimming channel.  Thus, $score_{match}$ can be used to identify a swimmer.

\subsection{Experimental Results}
\label{sec:4.4experimental_results}

We tested our mining algorithms on a set of $233$ videos (see Table~\ref{tab:swimmer_results}), showing over $130$ different athletes swimming in two structurally different swimming channels. Videos were recorded either at 720$\times$576@50i or at 1280$\times$720@50p. The videos cover different swimmers (in age, gender, physique, body size and posture) swimming in a swimming channel at different velocities between $1ms^{-1}$ and $1.75ms^{-1}$ and very different stroke rates. All mining was performed before any ground truth annotations were created.

\begin{table}
  \centering
  \caption{Swimming test video DB with mining results}
  \label{tab:swimmer_results}
  \begin{tabu}{llcccc} 
    \tabucline[1.5pt]{-}
    \multicolumn{2}{l}{Stroke}                                                                                                      & \multicolumn{1}{l}{Fly} & \multicolumn{1}{l}{Back} & \multicolumn{1}{l}{Breast} & \multicolumn{1}{l}{Free}  \\ 
    \tabucline[1.5pt]{-}
    \multicolumn{2}{l}{\# videos}                                                                                                   & 80                      & 28                       & 79                         & 46                        \\ 
    \hline
    \multirow{3}{*}{length~$[$s$]$}                                                                                       & min         & 18.3                    & 15.8                     & 19.3                       & 17.2                      \\
    & median      & 35.0                    & 31.2                     & 35.5                       & 33.9                      \\
    & max         & 72.7                    & 49.7                     & 85.7                       & 83.8                      \\ 
    \hline
    \multirow{3}{*}{\begin{tabular}[c]{@{}l@{}}GT stroke length \\$[$\# frames$]$\\\end{tabular}}                         & min         & 51                      & 58                       & 48                         & 52                        \\
    & median      & 67                      & 69                       & 69                         & 67                        \\
    & max         & 101                     & 85                       & 119                        & 108                       \\ 
    \hline
    \multirow{3}{*}{\begin{tabular}[c]{@{}l@{}}stroke length error\\$[$\# frames$]$~~ \end{tabular}}                      & avg         & 0.53                    & 0.32                     & 0.39                       & 0.39                      \\
    & \# 2        & 6                       & 0                        & 1                          & 0                         \\
    & \# not det. & 0                       & 1                        & 0                          & 1                         \\ 
    \hline
    \multicolumn{2}{l}{\# w/o det. stroke range}                                                                                    & 0                       & 2                        & 0                          & 0                         \\
    \hline
    \multicolumn{2}{l}{\begin{tabular}[c]{@{}l@{}}\% of detected cyclic~\\stroke range \end{tabular}}                               & 96.0                    & 84.5                     & 91.1                       & 82.8                      \\
    \hline
    \multicolumn{2}{l}{\begin{tabular}[c]{@{}l@{}}\% of erroneously detected~\\non-cyclic stroke range \end{tabular}}               & 1.8                     & 3.2                      & 6.0                        & 0.3                       \\
    \tabucline[1.5pt]{-}
  \end{tabu}
\end{table}

\begin{figure}[tb]
  \centering
  \includegraphics[width=0.98\columnwidth]{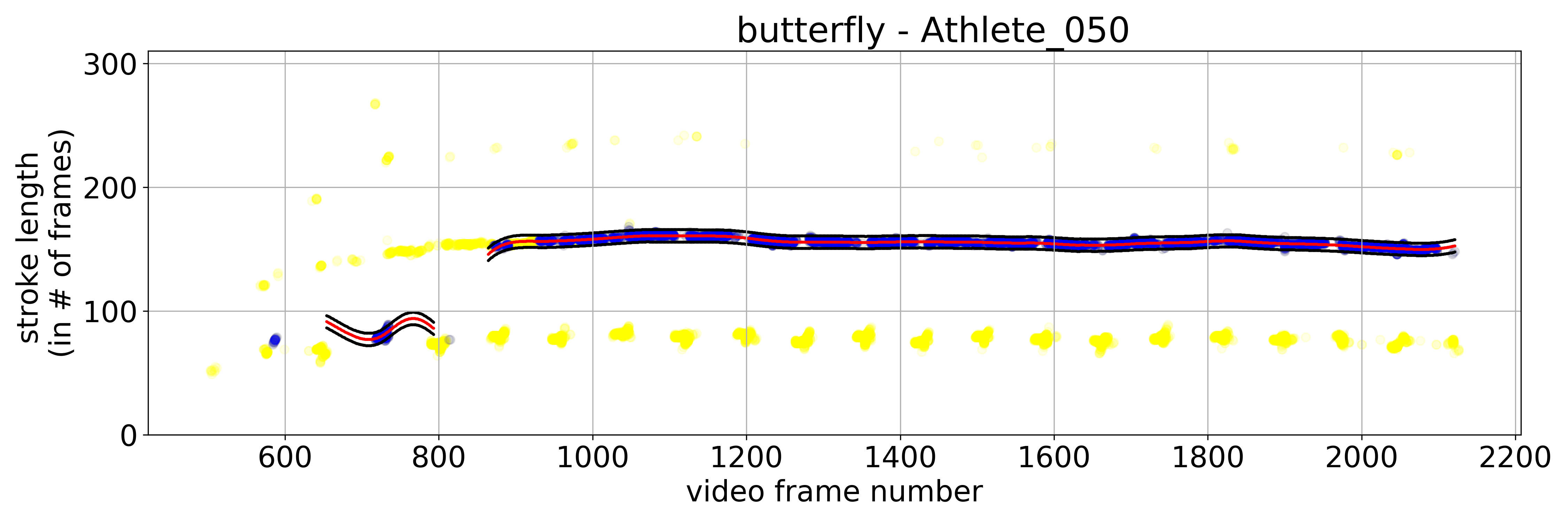}
  \caption{One of the 6 videos where the stroke length was incorrectly estimated twice as high as it actually was.}
  \label{fig:stroke_rate_failure}
\end{figure}

\textbf{Time-Continuous Cycle Speeds:} The precision of the time-continuous cycle speeds expressed by the number of frames per cycle was estimated by randomly picking one frame from each video and annotating it manually with the actual stroke length. In $2$ video sequences, our mining system did not determine a cycle speed at the frame of the ground truth. For another $6$ sequences the error in frames was larger than $2$, while for the remaining $225$ sequences the average deviation in frames from the ground truth was $0.43$ frames and $0.53$, $0.32$, $0.39$ and $0.39$ frames for breast, fly, back, and freestyle (see Table~\ref{tab:swimmer_results}). This exceptional quantitative performance can intuitively be grasped by a human observer from the stroke length graphs in Figure~\ref{fig:reoccurence_sequences}. In these graphs it is also visually striking if something has gone wrong, which was the case for $6$ videos. Figure~\ref{fig:stroke_rate_failure} depicts one of the few videos where the stroke length was incorrectly estimated twice as high as it actually was due to difficulties in detecting the joints reliably.

\textbf{Identify Cyclic Motion:} We annotated all $236$ videos roughly with the start and end time of the stroke. This sounds like an unambiguous task, but it was not: When the swimmer was starting the stroke out of the break-out from the dive, the starting point is fluent over some range. We decided to be more inclusive and marked the point early. However, it was extremely difficult to specify when the athlete stopped the stroke. Many athletes were drifting partially out of the image while still swimming when getting tired due to fast water velocities. This violated the assumption of our pose detection system that the simmer has to be completely visible. We decided to mark the end of the stroke range when a swimmer was knees downwards out of the picture. This choice, however, did not fit breast stroke well: During a cycle the swimmer pulls the heels towards the buttocks, bringing the feet back into the image, providing the system suddenly with a complete pose. We can see this effect in Table~\ref{tab:swimmer_results}, there our algorithm over-detects up to $6\%$ of the stroke range according to our early cut-off ground truth. This over-detection is primarily an artifact of how we determined the ground truth range of the stroke, but no real error. Our mining algorithm detected overall $89.5\%$ of all ground truth stroke ranges, while only detecting $3.1\%$ additionally outside. This performance is more than sufficient in practice. Moreover, the length of the detected cyclic motion range(s) per video was an excellent indicator to identify unstable and/or erroneous pose detection results. A cyclic motion range of less than $10$ seconds indicated that our automatic pose detection system had difficulties to detect the human joints due to strong reflections, water splashes, spray and/or air bubbles in the water. For these sequences determining the stroke cycle stability based on the identified temporally striking poses of the athlete does not make sense. Hence, in the subsequent experiments, only cyclic motion sequences of $10$ seconds or longer were used. This reduced the number of videos from $233$ down to $213$.

\textbf{Temporally Striking Poses:} Poses which are temporally salient and unambiguously easy to determine by humans typically focus on one or two characteristic angles. An example is when the upper arm is vertical in freestyle (in the water) or backstroke (outside the water). Everything else of the pose is ignored. This is not how our temporally striking pose is defined: a pose which is easy to localize temporally by our system. Due to this mismatch between what the human is good at and our system, we only evaluate the temporally striking poses indirectly via their use to capture cycle stability.

\textbf{Cycle stability:} For each video we computed the stroke stability indicator value based on a single reference stroke clip. The reference stroke clip was selected by using the ground truth frame from the time-continuous cycle speed evaluation as the end point and by subtracting our estimated stroke length from that to compute the start frame. For each stroke we sorted the videos based on its stroke cycle stability indicator value and picked randomly one video from the top $20\%$, one from the middle $20\%$ and one from the bottom $20\%$. We then asked a swim coach to sort these three videos based on his assessed stroke cycle stability. We compared the result to the automatically computed ordering. Very similar results were obtained with the temporally striking poses as reference:

\textit{Breast:} There was an agreement in the ordering of the videos ranked 1st and 2nd. The athlete of the first video showed off an exceptionally stable stroke pattern. However, the video ranked 3rd was judged by the coach as being equivalent to the one ranked 2nd. The 3rd video is one of the instances there the swimmer is getting tired, drifting regularly with his lower legs out of the picture during the stretching phase in breast stroke. This explains the discrepancy between the judgement of the coach and our system.

\textit{Fly:} The coach and the system agreed on the ordering. We also notice that our system was picking up those athlete, who were breathing every other stroke and exhibit a strong difference between the cycle with and without the breath. With respect to a two-cycle pattern their stroke was stable. Typically, coaches emphasize that there should be as little difference as possible between a breathing cycle and a non-breathing cycle.

\textit{Back:} The coach and the system agreed on the ordering.

\textit{Free:} The coach was ranking the second video as having a slightly better stroke stability than the first video. They agreed on the video ranked 3rd as the athlete was showing an unsteady and irregular flutter flick. The discrepancy between the first two videos can be explained by peculiarities of the video ranked 2nd: water flow speed was higher than normal, leading to a slightly higher error frequency in the automatically detected poses.


\section{Mining Long Jump Pose Data}
\label{sec:5}

\begin{figure*}[tb]
  \centering
  \includegraphics[width=0.98\textwidth]{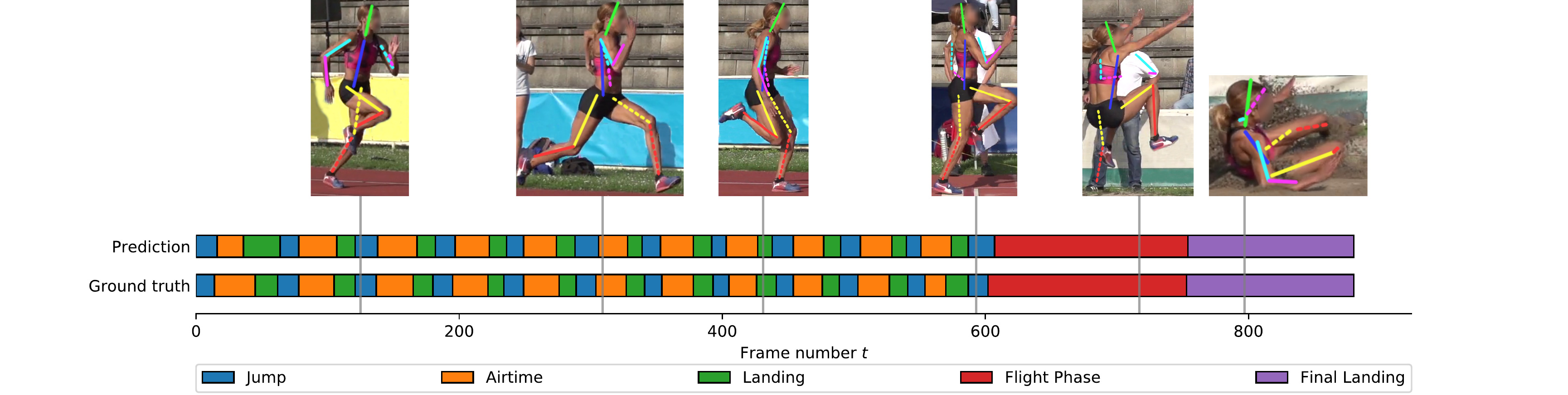}
  \caption{Qualitative comparison of predicted and ground truth long jump phases in one test video. Exemplary video frames and their estimated poses are depicted for each phase.}
  \label{fig:long_jump_example}
\end{figure*}

As a second example for pose data mining, we look at data of long jump athletes recorded at athletics championships and training events. Long jumping is different from swimming in many respects: Firstly, long jump features only semi-cyclic movement patterns. While the run-up is composed of repetitive running motion, the final jump itself is strikingly different and only performed once per trial. Secondly, the action is performed over a complete running track and recorded by a movable camera from varying angles. Third, spectators and other objects in the background along the track are likely to cause regular false detections of body joints. Our data consists of $65$ videos recorded at $200$Hz, where each video shows one athlete during a long jump trial from the side. The camera is mounted on a tripod and panned from left to right to track the athlete. The videos cover various athletes and six different long jump tracks. Figure ~\ref{fig:long_jump_example} shows exemplary video frames from one trial. The long jump pose database consists of $45,436$ frames with full-body pose estimates.

\subsection{Automatic Temporal Classification of Long Jump Pose Sequences}
\label{sec:5.1long_jump_sequence_classification}

Video based performance analysis for long jump athletes involves various time dependent measures like the number of steps until the final jump, the relative joint angles during the run-up, the vertical velocity during the final jump, and the flight phase duration. To obtain such measures automatically, pose information alone does not suffice. Instead it requires to pick the poses from the right phase of a long jump. Therefore, we present here how to mine the pose data to temporally identify the different phases of a long jump such that the phase specific performance measures can be computed from the detected poses. We partition a long jump action during one trial into a periodic and an aperiodic part. The periodic run-up consists of repeated \textit{jumps} (the rear leg pushes the body upwards), \textit{airtimes} (no contact with the ground) and \textit{landings} (from first contact with the ground till the jump phase). The aperiodic part consists of the \textit{flight phase} and the \textit{final landing} in the sandpit. We annotated the long jump videos with respect to these five phases. Given a long jump video of length $T$ and the extracted pose sequence $p_{1:T}$, our mining task is now to predict the phase class $c_t \in C = \{ \text{jump}, \text{airtime}, \dotsc, \text{final landing} \}$ the athlete is in at each time step $t \in [1,T]$. Figure~\ref{fig:long_jump_example} depicts exemplary frames for each phase.

\textbf{Pose Clustering:} Since the pose space itself is large, finding a direct mapping from the pose space to the possible long jump phases $C$ is difficult. Similar to the cyclic strokes in swimming we expect poses in identical long jump phases to be similar to each other. We expect this to be true even across videos of different athletes and slightly varying camera viewpoints. This leads to assumption 1: \textit{Similar poses often belong to the same phase} (Asm. 1).

Instead of learning a direct mapping from pose to phase, we first partition the space of poses into a fixed number of subspaces. Henceforth, each pose is described by the discrete index of its subspace. As long as the subspace partition preserves similarity, we expect that the distribution of phases in one pose subspace is informative, i.e. non uniform with respect to phase class $c_t$. Let $S$ be the set of poses in our database. We perform unsupervised $k$-Medoids clustering on $S$ with our normalized pose similarity measure from Equation~(\ref{eq:mse_norm}) to create our subspace partition. The clustering defines a function $h(p) \mapsto [1,k]$ that maps a pose $p$ to the index of its nearest cluster centroid. With Asm. 1 we define the probability $P(c \vert h(p))$ as the fraction of poses in cluster $h(p)$ labeled with phase $c$:
\begin{equation}
  \label{eq:phase_given_pose}
  P( c \vert h(p) ) = \frac{\left\vert \left\{ p_i \in S \middle\vert h(p_i) = h(p) \wedge c_i = c \right\} \right\vert}{\left\vert \left\{ p_i \in S \middle\vert h (p_i) = h(p) \right\} \right\vert}
\end{equation}

\textbf{Markov Representation of Long Jump Sequence:} With Equation~(\ref{eq:phase_given_pose}) we could already predict the phase for each pose in a video individually. However, noisy predictions and phase-unspecific poses may render Asm. 1 in a fraction of the poses as incorrect. We have to incorporate the complete pose sequence to obtain correct phase predictions even for frames with wrongly estimated or ambiguous poses. With the rigid long jump movement pattern and the chosen phase definition, we can make two more assumptions: \textit{An athlete stays in a phase for some time before entering a different phase. Subsequent poses are likely to belong to the same phase} (Asm. 2).\textit{ Also, the possible transitions between long jump phases are limited by a fixed sequential pattern} (Asm. 3).

We can model these assumptions by stating the temporal succession of long jump phases as a state transition graph. Each state corresponds to one possible phase. Asm. 2 and 3 are reflected by self-loops and a small number of outgoing edges at each state, respectively. At each time step $t$ the athlete is in a phase which we cannot directly observe. The pose (or rather its estimate) at time $t$ is observable, however. Combining the graph with emission probabilities $P(h(p) \vert c)$ and transition probabilities $P(c_{t+1} \vert c_t)$ we obtain a classical Hidden Markov Model. The emission probabilities $P(h(p) \vert c)$ can be computed as
\begin{equation}
  \label{eq:emission_prob}
  P( h (p) \vert c ) = \alpha \cdot P ( c \vert h(p)) \cdot P ( h(p) ),
\end{equation}
where $\alpha$ is a normalization constant. The transition probabilities are obtained similarly by counting the number of observed transitions in the dataset.

Given a new long jump video and the corresponding pose sequence $p_{1:T}$ we first transform the sequence to the clustering-based discrete pose description $h(p_{1:T})$. We then use the Viterbi algorithm for the most likely phase sequence $c_{1:T}^{*}$ with
\begin{equation}
  \label{eq:viterbi_phase_sequence}
  c_{1:T}^{*} = \arg \max_{c_{1:T}} P \left( c_{1:T}    \middle\vert h(p)_{1:T} \right).
\end{equation}


\subsection{Experimental Results}
\label{sec:5.2experimental_results}

\begin{table}
  \caption{Results of long jump phase detection (AP) with IoU threshold $\tau=0.5$ (upper part) and the derived length and step count during the long jump run-up (lower part).}
  \label{tab:long_jump_results}
  \centering
  \begin{tabu}{lc|lc} 
    \tabucline[1.5pt]{-}
    Jump                                                                                                                        & 0.84                 & Flight Phase                    & 0.94  \\
    Airtime                                                                                                                     & 0.91                 & Final Landing                   & 0.97  \\
    Landing                                                                                                                     & 0.80                 & \multicolumn{2}{l}{}                    \\ 
    \hline
    mAP                                                                                                                         & \multicolumn{1}{c}{} &                                 & 0.89  \\ 
    \tabucline[1.5pt]{-}
    \multicolumn{2}{l}{\multirow{3}{*}{\begin{tabular}[c]{@{}l@{}}\# videos with given abs.\\error in step count\end{tabular}}}                        & $\vert error_{steps} \vert = 0$ & 53    \\
    \multicolumn{2}{l}{}                                                                                                                               & $\vert error_{steps} \vert = 1$ & 7     \\
    \multicolumn{2}{l}{}                                                                                                                               & $\vert error_{steps} \vert > 1$ & 0     \\ 
    \hline
    \multicolumn{3}{l}{\begin{tabular}[c]{@{}l@{}}Average abs. error in derived\\run-up length [s]\end{tabular}}                                                                         & 0.06  \\
    \tabucline[1.5pt]{-}
  \end{tabu}
\end{table}

Although we formulated our problem as a per-frame classification task, the predictions should reflect the sequential phase transitions as well as the length of each annotated phase. Therefore, we evaluate our phase detection mining by the standard protocol of average precision (AP) and mAP for temporal event detection in videos \cite{Gorban2015, Heilbron2015}. For each video we combine sequential timestamps belonging to the same long jump phase $c$ into one \textit{event} $e_j=(t_{j,1}, t_{j,2}, c_j)$ with $t_{j,1}$ and $t_{j,2}$ being the start and stop time of the event. Let $E=\{ e_j \}_{j=1}^{J}$ be the set of sequential events in one video. In the same manner we split the predicted phase sequence $c_{1:T}^{*}$ into disjoint predicted events $e_j^*$. Two events match temporally if their intersection over union (IoU) surpasses a fixed threshold $\tau$. A predicted event $e_j^*$ is correct if there exists a matching ground truth event $e_j \in E$ in the same video with
\begin{equation}
  \label{eq:phase_sequence_match}
  c_j = c_j^* \wedge \frac{\left[ t_{j,1}, t_{j,2} \right] \cap \left[ t_{j,1}^*, t_{j,2}^* \right]}{\left[ t_{j,1}, t_{j,2} \right] \cup \left[ t_{j,1}^*, t_{j,2}^* \right]} > \tau.
\end{equation}
We optimize clustering parameters on a held-out validation set and use the remaining $60$ videos to evaluate our approach using six-fold cross evaluation. Table~\ref{tab:long_jump_results} depicts the results at a fixed $\tau=0.5$ IoU threshold. We achieve a mAP of $0.89$ for long jump phase detection. Due to their length and the unique poses observed during the flight and landing in the sandpit, these two phases are recognized very reliably with $0.94$ and $0.97$ AP, respectively. The phases of the periodic part show more uncertainty since each phase is considerably shorter and poses of the jump-airtime-landing cycle are more similar to each other.Figure~\ref{fig:long_jump_example} depicts qualitative results on one test video. Our method is able to reliably divide the cyclic run-up and the final flight phase and landing. Few predictions for the periodic phases are slightly misaligned, but the overall cyclic pattern is preserved. The phase predictions can directly be used to derive further kinematic parameters like the duration of the run-up and the number of steps. The results in Table~\ref{tab:long_jump_results} show that the run-up duration can be derived very accurately with an average deviation of $60$ms. The correct number of steps is recovered in the majority of videos.

\section{Conclusion}
\label{sec:6}
Noisy pose data of individual sport recordings will soon be available in abundance due to DNN-based pose detections systems. This work has presented unsupervised mining algorithms that can extract time-continuous cycle speeds, cycle stability scores and temporal cyclic motion durations from pose sequences of sport dominated by cyclic motion patterns such as swimming. We also showed how to match pose clips across videos and identify temporally striking poses. As it has become apparent from the analysis, results from our mining algorithms can be further improved if automatic pose detection system focus on dealing with athletes that are not fully visible in the video. We additionally apply our concept of pose similarity to pose estimates in long jump recordings. We model the rigid sequential progression of movement phases as a Markov sequence and combine it with an unsupervised clustering-based pose discretization to automatically divide each video into its characteristic parts. We are even able to identify short intra-cyclic phases reliably. The derived kinematic parameters show a direct application of this approach.

\section*{Acknowledgement}
This research was partially supported by FXPAL during Rainer Lienhart's sabbatical. He thanks the many colleagues from FXPAL (Lynn Wilcox, Mitesh Patel, Andreas Girgensohn, Yan-Ying Chen, Tony Dunnigan, Chidansh Bhatt, Qiong Liu, Matthew Lee and many more) who greatly assisted the research by providing an open-minded and inspiring research environment.

\bibliographystyle{ACM-Reference-Format}
\bibliography{bibliography}

\end{document}